\documentclass{article}
\usepackage{spconf,amsmath,graphicx}
\usepackage{multirow}
\usepackage{xspace}
\usepackage{balance}
\usepackage{color}

\title{CC-Net: Image Complexity Guided Network Compression for Biomedical Image Segmentation}
\name{Suraj Mishra, Peixian Liang, Adam Czajka, Danny Z. Chen, X. Sharon Hu\thanks{Copyright \textcopyright 2018 IEEE}}
\address{Dept. of Computer Science and Engineering, University of Notre Dame, Notre Dame, IN 46556, USA}
\begin{document}
%
\maketitle
\begin{abstract}
Convolutional neural networks (CNNs) for biomedical image analysis are often of very large size, resulting in high memory requirement and high latency of operations. Searching for an acceptable compressed representation of the base CNN for a specific imaging application typically involves a series of time-consuming training/validation experiments to achieve a good compromise between network size and accuracy. To address this challenge, we propose CC-Net, a new image complexity-guided CNN compression scheme for biomedical image segmentation. Given a CNN model, CC-Net predicts the final accuracy of networks of different sizes based on the average image complexity computed from the training data. It then selects a multiplicative factor for producing a desired network with acceptable network accuracy and size. Experiments show that CC-Net is effective for generating compressed segmentation networks, retaining up to $\approx 95\%$ of the base network segmentation accuracy and utilizing only $\approx 0.1\%$ of trainable parameters of the full-sized networks in the best case. 
\end{abstract}
\begin{keywords}
Biomedical image segmentation, Deep neural networks, Network compression, Image complexity
\end{keywords}
\section{Introduction}
\label{sec:intro}
CNNs are often of very large size, resulting in high memory requirement and high latency of operations, and thus not suitable for resource-constrained applications (e.g., edge computing). To find a good compromise between network size and performance, a series of time-consuming training/validation experiments is often used for a specific imaging application. To address this challenge, we propose a new network compression scheme targeting biomedical image segmentation in resource-constrained application settings (e.g., low cost and easy-to-carry imaging devices for disaster/emergency response and military rescue).

Since the inception of FCNs \cite{fcn}, various improved segmentation networks \cite{unet,cumednet,Suggestive,cascaded} were developed. To compress CNNs, various pre-training \cite{squeezenet,mobilenet} and post-training compression \cite{deepcompression,nvidiapruning} schemes were suggested. In these techniques, compression thresholds often need to be set manually in multiple pruning iterations. 

In contrast with natural scene images, in biomedical or healthcare application settings, images are often for a specific type of disease/injury and captured by specific imaging devices; hence, their objects and settings are quite ``stable", making the image characteristics and complexity much more specific to analyze. In this paper, we leverage this observation to introduce CC-Net.

Based on the image complexity measure, target CNN, and user constraints (e.g., desired accuracy or available memory), CC-Net determines  for the given dataset the most suitable multiplicative factor to compress the original CNN. The resulting compressed network is then trained, with much less effort and memory compared to the original network. Experiments using 5 public and 2 in-house datasets and 3 commonly-used CNN segmentation models as representative networks show that CC-Net is effective for compressing segmentation networks, retaining up to $\approx 95\%$ of the base network segmentation accuracy and utilizing only $\approx 0.1\%$ of trainable parameters of the full-sized networks in the best case. 

\section{Methodology}
\label{sec:method}

\begin{figure}[tb]
    \includegraphics[width=0.48\textwidth]{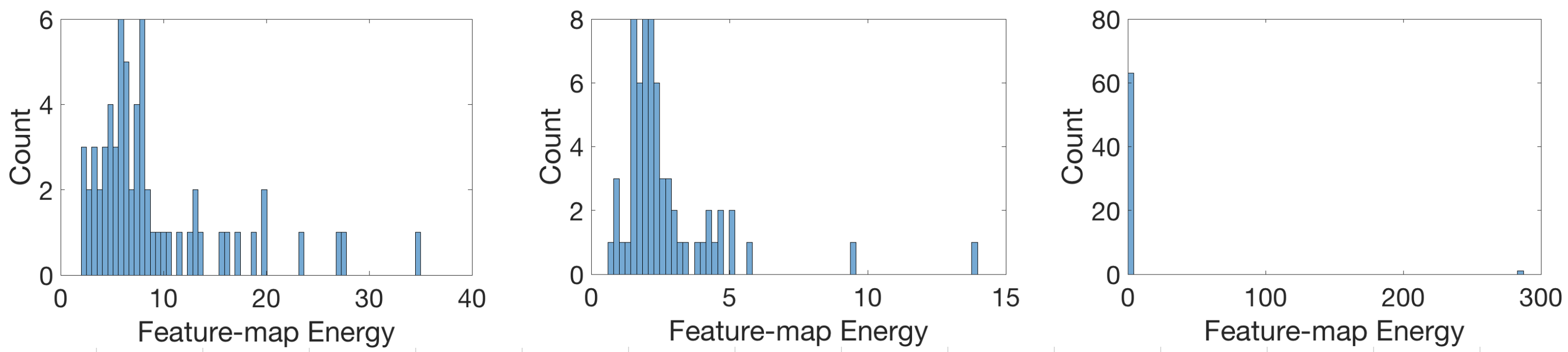}
    \centering
    \caption{Feature map energy distributions of the first convolutional layer of U-Net for several datasets: (left) gland images (high complexity), (middle) C2DH-HeLa cell images, and (right) wing-disk images (low complexity).}
    \label{fig:fm_distribution}  
        \vspace*{-0.18in}
\end{figure}

Feature-map (filter output) energy is a good indicator of filter's feature extraction capability. We have conducted a large set of experiments to study the relationship between feature-map energy and training datasets.  Fig.~\ref{fig:fm_distribution} depicts 3 example energy distribution for the first convolution layer of U-Net~\cite{unet}. One can observe that (i) a significant number of filter outputs have very low energy, and (ii) less ``complex'' (to be defined more precisely later) datasets have more low-energy filter outputs. These suggest that U-Net \cite{unet} may be unnecessarily large for some biomedical datasets, and in these cases, filters can be pruned without significantly deteriorating the accuracy.

\begin{figure}[tb]
    \includegraphics[width=0.48\textwidth]{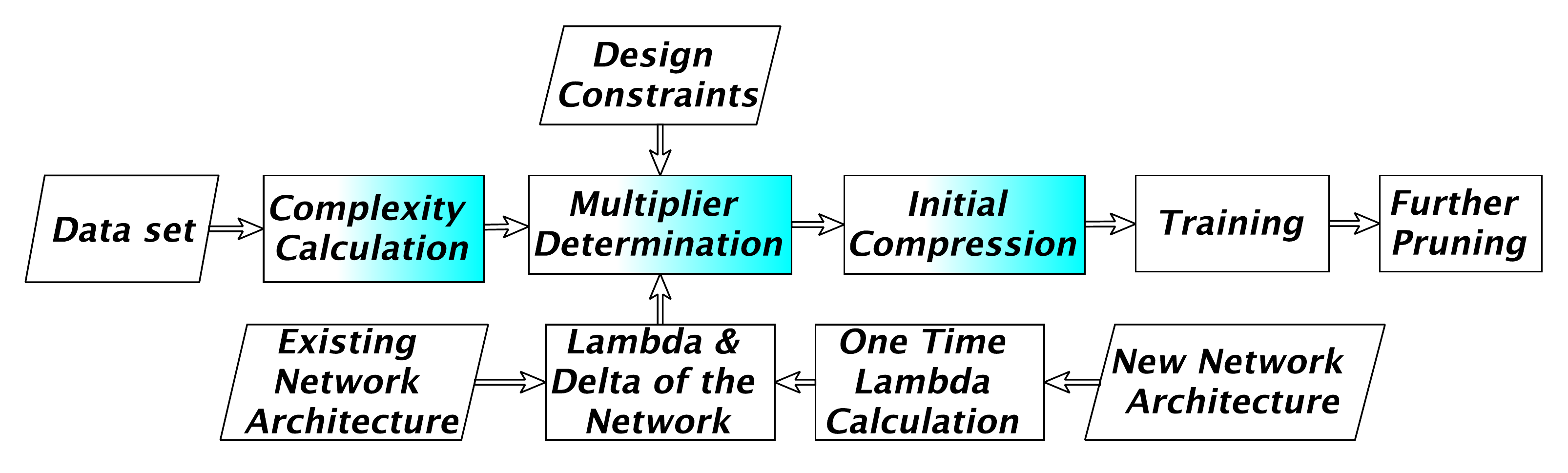}
    \centering
    \caption{Our proposed scheme for CC-Net.}
    \label{fig:pipeline}    
        \vspace*{-0.2in}
\end{figure}

Based on above observations, we develop CC-Net, depicted in Fig.~\ref{fig:pipeline}. Inputs and internal operations of CC-Net are shown in parallelograms and rectangles. Existing architectures are the 3 CNNs studied and parameterized in our work. Colored boxes highlights the key contributions of this paper. We elaborate the major components in CC-Net below.

\subsection{Image Complexity Computation}
\label{sec:complexity}

We seek an image complexity metric that can (i) indicate the trends of segmentation accuracy and (ii) be easily computed. Our work examined the following candidate metrics: (i) signal energy, (ii) edge information (Sobel and Scharr filters along with image pyramid), (iii) local key-point detection using SURF \cite{surf}, (iv) visual clutter information \cite{visualclutter}, (v) JPEG complexity \cite{imagecomplexity} and (vi) blob density. To obtain a single complexity value for an entire dataset, we take the average of complexity values over all the images in the dataset.

\begin{table}[thb!]
\caption{Datasets and properties.}
\scriptsize
\begin{center}
\begin{tabular}{ p{2.0cm} p{0.2cm} p{0.8cm} p{0.5cm} p{0.5cm} p{0.9cm}  }
 \hline
 \multicolumn{0}{c}{\rule{0pt}{6pt}
                   Dataset} & \multicolumn{0}{c}{Size} & \multicolumn{0}{c}{Type} & \multicolumn{0}{c}{J} & \multicolumn{0}{c}{B} & \multicolumn{0}{c}{Source} \\
\hline\rule{0pt}{6pt}
Glands (GL)  & 165 & RGB & 0.2401 & 0.5711 & \cite{gland} \\
Lymph Nodes (LN)  &  74 & Ultrasound & 0.2445 & 0.0715 & in-house\\
Melanoma (ME)  &  2750 & RGB & 0.1505 & 0.3055 & \cite{melanoma} \\
C2DH-HeLa (CH)  & 20 & Gray & 0.1403 & 0.4607 & \cite{dataset} \\
Wing Discs (WD)  & 996 & Gray & 0.0925 & 0.1348 & in-house \\
C2DH-U373 (CU)  & 34 & Gray & 0.1473 & 0.0699 & \cite{dataset}\\
C2DL-PSC (CP)  & 4 & Gray & 0.2296 & 0.3066 & \cite{dataset} \\
 \hline
\end{tabular}
\end{center}
\label{tab:data}
    \vspace*{-0.2in}
\end{table}

\begin{figure*}[tb]
    \centering
    \includegraphics[width=0.70\textwidth]{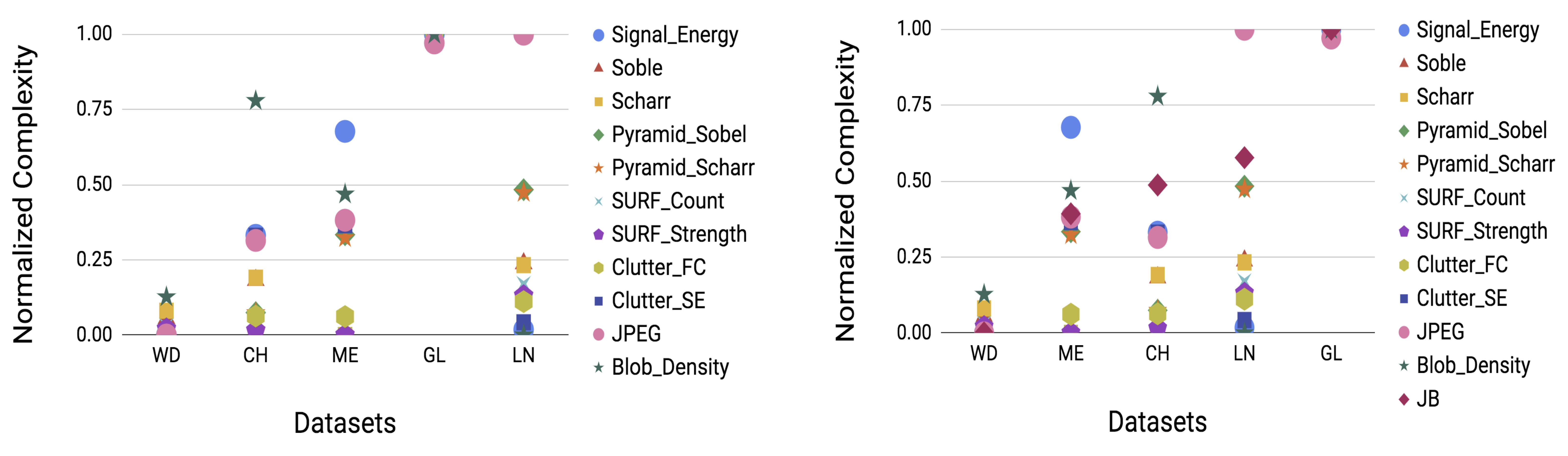}
         \vspace*{-0.1in}
    \caption{Mapping image complexity with accuracy degradation. (left) Our datasets are arranged in increasing order of drop in F1 score; (right) our datasets are arranged in increasing order of drop in IU score.}
    \label{fig:complexity_trend}
       \vspace*{-0.15in}
\end{figure*}

Out of 7 datasets shown in Table.~\ref{tab:data}, 5 datasets (train-set, top 5 rows) are used to formulate the methodology, while the remaining 2 datasets (test-set) are used for blind evaluation. Fig.~\ref{fig:complexity_trend} plots average complexities (normalized to the range [0,1]) against the train-set datasets arranged as their  F1 and IU score degradation (two most popular segmentation accuracy metrics). Among these complexity measures, the JPEG complexity better follows the trend of F1 score degradation (i.e., higher complexity leads to lower F1). Since IU is related to both feature variety and quantity, to represent it,  we linearly combine the JPEG complexity $J$ and blob density $B$ ($B = \sum_{i} fg\_pixel / \sum_{i} img\_pixel$, see Table \ref{tab:data}), as $JB = \omega J + (1 - \omega) B$, where 
$\omega$ is a value in $[0,1]$. The value of $\omega$ is determined by inspecting the optimal regression fitting on the training datasets in our experiments. We consider J and JB for multiplier determination explained as follows.

\subsection{Multiplier Determination and Network Compression}
\label{sec:multiplier}
Keeping all other variables unchanged, we can express the relationship between the segmentation accuracy ($A$) and data complexity ($C$) as $A = f(\theta, C)$, where  $\theta$ is the number of trainable parameters in a CNN. For general networks, the function $f(\theta,C)$ can be rather complicate. But in general, segmentation accuracy is monotonically non-decreasing with respect to $\theta$ and $C$, i.e., $\frac{\partial f}{\partial \theta} \geq 0$ and $\frac{\partial f}{\partial C} \geq 0$. 

\begin{figure}[tb]
    \centering
    \includegraphics[width=0.48\textwidth]{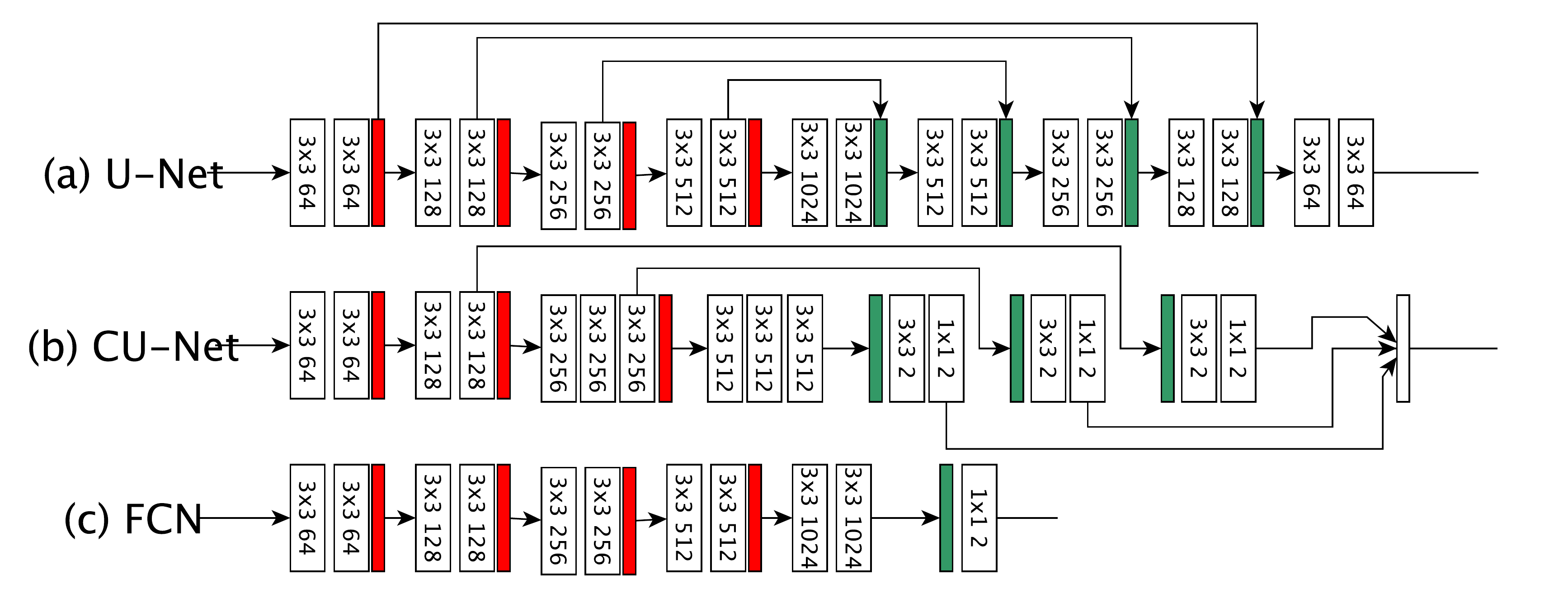}
    \caption{CNN architectures. A colorless block represents a group of convolution, batch normalization, and ReLU. A red block and a green block represent pooling and up-scaling operations, respectively.}
    \label{fig:network}
        \vspace*{-0.2in}
\end{figure}

For CNNs (see Fig.~\ref{fig:network}), we observe (as discussed in Section \ref{sec:exp}) that $\frac{\partial f}{\partial \log \theta}$ can be approximated by a linear function of $C$. That is, $\frac{\partial f}{\partial \log \theta} \approx \lambda C + \delta$ for a constant $\lambda$ that reflects the {\it degree of degradation}. Given the linear dependency, if $\lambda$ and $\delta$ are known, then it is straightforward to compute the change in accuracy or in the number of parameters, when the other is provided. The value of $\lambda$ is network-dependent, and can be obtained by performing systematic analysis on network compression and tracking the change in accuracy.

A simple way of compression is to uniformly scale down the number of feature maps in every convolution layer using a single multiplier ($\alpha \in (0,1]$). Existing work has shown that it performs very well~\cite{mobilenet,morphnet}. The number of trainable parameters after scaling becomes $\theta^* = \alpha FM_i$ $\times F_i^X \times F_i^Y \times \alpha FM_{i+1}$, where $FM_i$ and $FM_{i+1}$ are the numbers of input and output feature maps, and $F_i^X$ and $F_i^Y$ are filter dimensions. However, finding a good $\alpha$ is challenging. We employ complexity measures to determine $\alpha$.

When producing compressed networks, we consider two practical scenarios: \textbf{(1)} memory-constrained best possible accuracy, and \textbf{(2)} accuracy-guided least memory usage. For (1), two sub-cases are: (1.a) disk space budget and (1.b) main memory budget. 
For case (1.a), given a disk space budget in MB, we first determine $\theta^*$, based on the number of bits for each parameter. Then $\alpha$ can be computed as $\alpha = \sqrt{\frac{\theta^*}{\theta}}$. For case (1.b), sizes of feature-maps are considered along with the number of bits for $\theta^*$, and the value of $\alpha$ can be determined as $\alpha = \frac{\theta^*}{\theta}$. For (2), given the lowest acceptable accuracy $A_{min}$ and the original base network accuracy $A_{org}$, using the linear model, $A_{org}-A_{min} = (\lambda C + \delta)(\log \theta - \log \theta^*)$, $\theta^*$ and so as $\alpha$ can be readily computed. Using $\alpha$, a compressed network is produced, which then can be trained.

\begin{figure*}[tb]
  \centering
  \centerline{\includegraphics[width=0.76\textwidth]{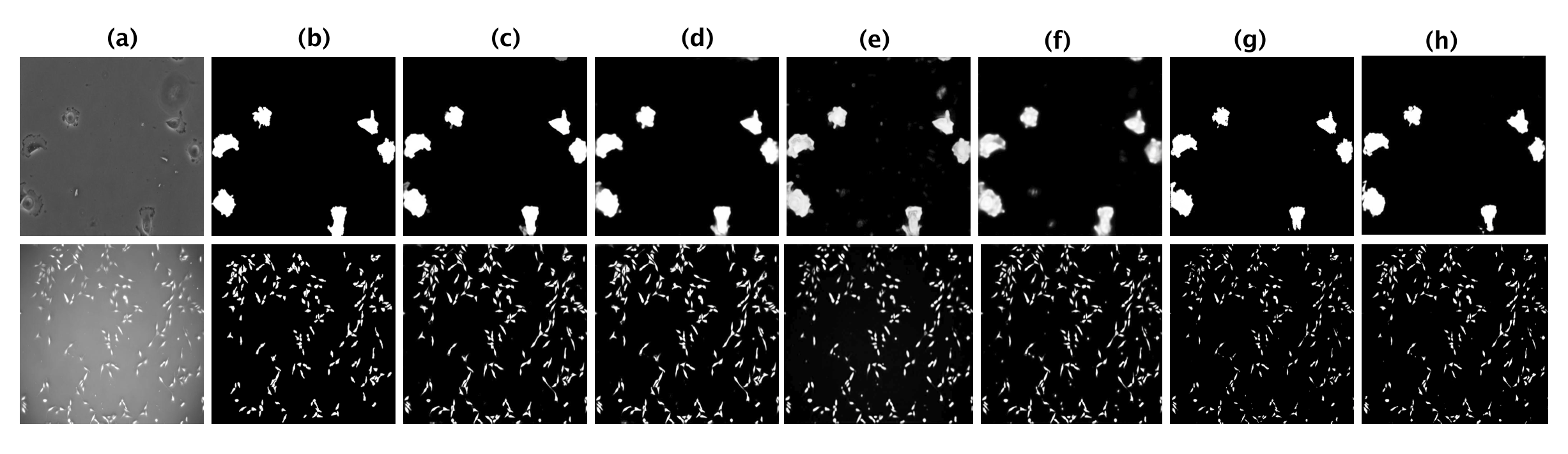}}
    \vspace*{-0.1in}
\caption{Some C2DH-U373 (top row) \& C2DL-PSC (bottom row) segmentation output. (a) Input images, (b) ground truth, the segmentation output of (c) U-net, (d) CC-U-Net, (e) U-Net + \cite{nvidiapruning}, (f) CC-U-Net + \cite{nvidiapruning}, (g) U-Net + \cite{squeezenet}, and (h) CC-U-Net + \cite{squeezenet}.}
\label{fig:segmentation}
    \vspace*{-0.2in}
\end{figure*}

\section{Experimental Evaluation}
\label{sec:exp}
\begin{figure}[tb]
    \centering
    \includegraphics[width=0.48\textwidth]{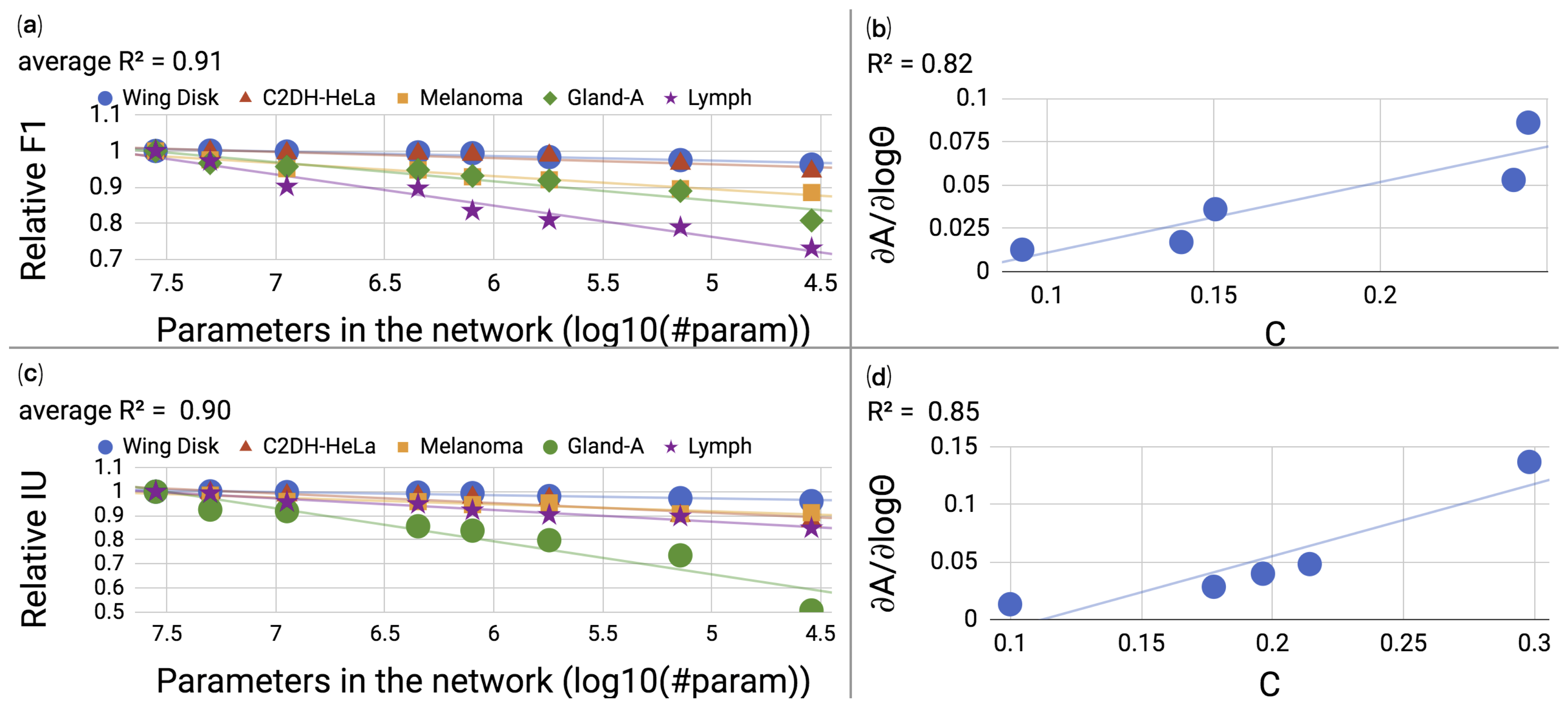}
    \caption{Calculated degree of degradation ($\lambda$) for the FCN architecture. F1 ($\lambda_{F1} = 0.407$ and $\delta_{F1} = -0.030$) and IU ($\lambda_{IU} = 0.627$ and $\delta_{IU} = -0.070$).}
    \label{fig:fcn}
        \vspace*{-0.2in}
\end{figure}

\begin{figure*}[tb]
    \centering
    \includegraphics[width=0.89\textwidth]{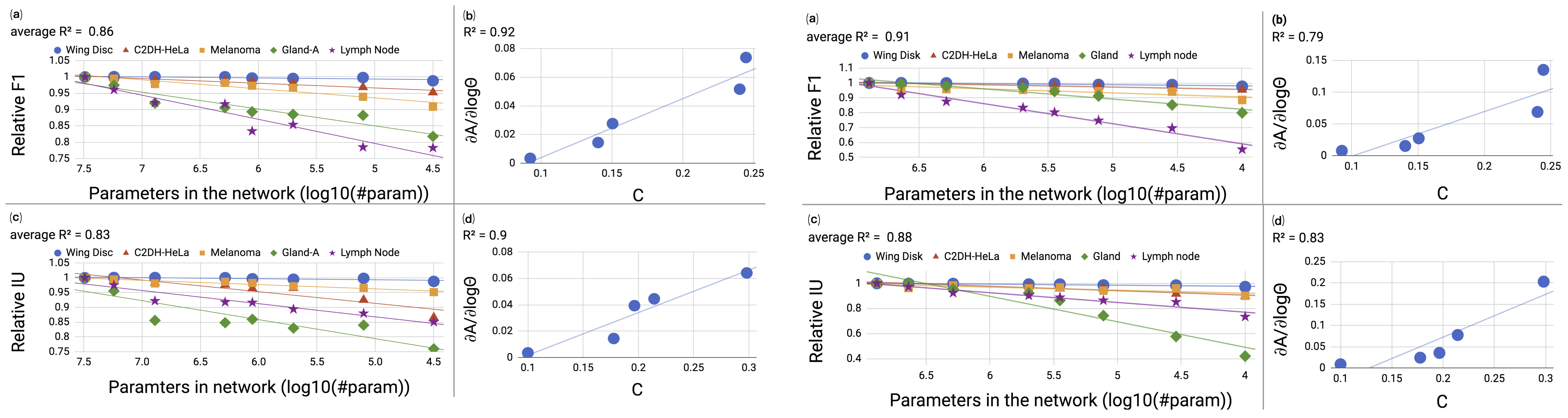}
     \vspace*{-0.05in}
    \caption{Calculated degree of degradation ($\lambda$). U-Net (left): F1 ($\lambda_{F1} = 0.411$ and $\delta_{F1} = -0.037$) and IU ($\lambda_{IU} = 0.323$ and $\delta_{IU} = -0.031$); CUMedVision (right): F1 ($\lambda_{F1} = 0.701$ and $\delta_{F1} = -0.071$) and IU ($\lambda_{IU} = 1.010$ and $\delta_{IU} = -0.129$).}
    \label{fig:res}
        \vspace*{-0.2in}
\end{figure*}

5 train-set datasets (Glands, Lymph Nodes, Melanoma, C2DH-HeLa, Wing Discs) are used to determine $\frac{\partial A}{\partial \log \theta}$ for 3 CNN models (Fig.~\ref{fig:network}), which is then mapped to J \& JB to determine $\lambda$. For simple calculations maintaining integer filter values, $\alpha \in \{1,0.75,0.5,0.25$, $0.1875$, $0.125,0.0625,0.03125\}$, are considered (Fig.~\ref{fig:fcn} \& Fig.~\ref{fig:res} (a), (c) X-axis). 2 test-set datasets (C2DH-U373, C2DL-PSC) are used to validate our method. We use a standard back-propagation implementing Adam (learning rate = 0.00005) and cross entropy as loss function using data augmentation. Experiments are performed on NVIDIA-TITAN and Tesla P100 GPUs, using the Torch framework.

Fig.~\ref{fig:segmentation} shows some segmentation output. Fig.~\ref{fig:fcn} and \ref{fig:res} show the calculated degree of degradation ($\lambda$) for FCN \cite{fcn}, U-Net \cite{unet}, and CUMedVision \cite{cumednet} networks. In these figures, (a) and (c) give the degradation in the relative F1 and IU accuracy (i.e., $\frac{Acc_\alpha}{Acc_{\alpha=1}}$) with respect to changes in the number of parameters expressed in logarithmic values. The slopes of regression lines for each dataset in (a) and (c) are plotted against the respective complexities in (b) and (d).

\begin{figure}[tb]
  \centering
 \includegraphics[width=0.45\textwidth]{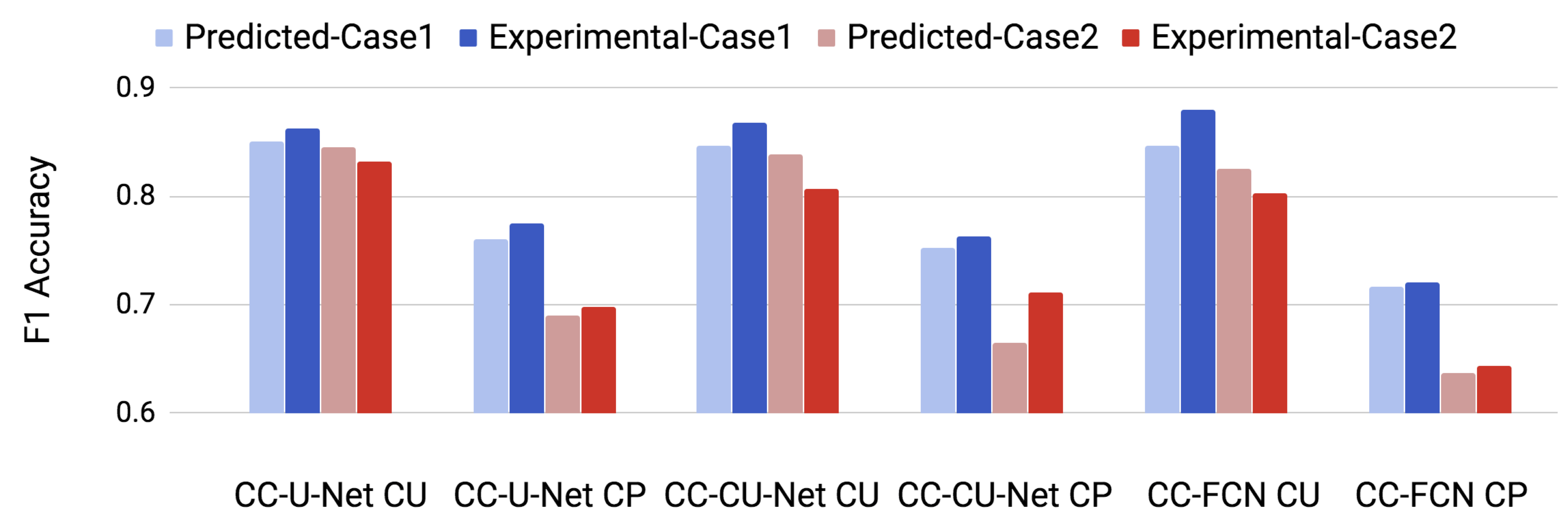}
    \vspace*{-0.05in}
\caption{Predicted (CC-Net) and experimental F1 scores for test cases 1 (accuracy constraint) and 2 (memory constraint).}
\label{fig:effectiveness}
   \vspace*{-0.1in}
\end{figure}

\begin{figure}[tb]
  \centering
 \includegraphics[width=0.48\textwidth]{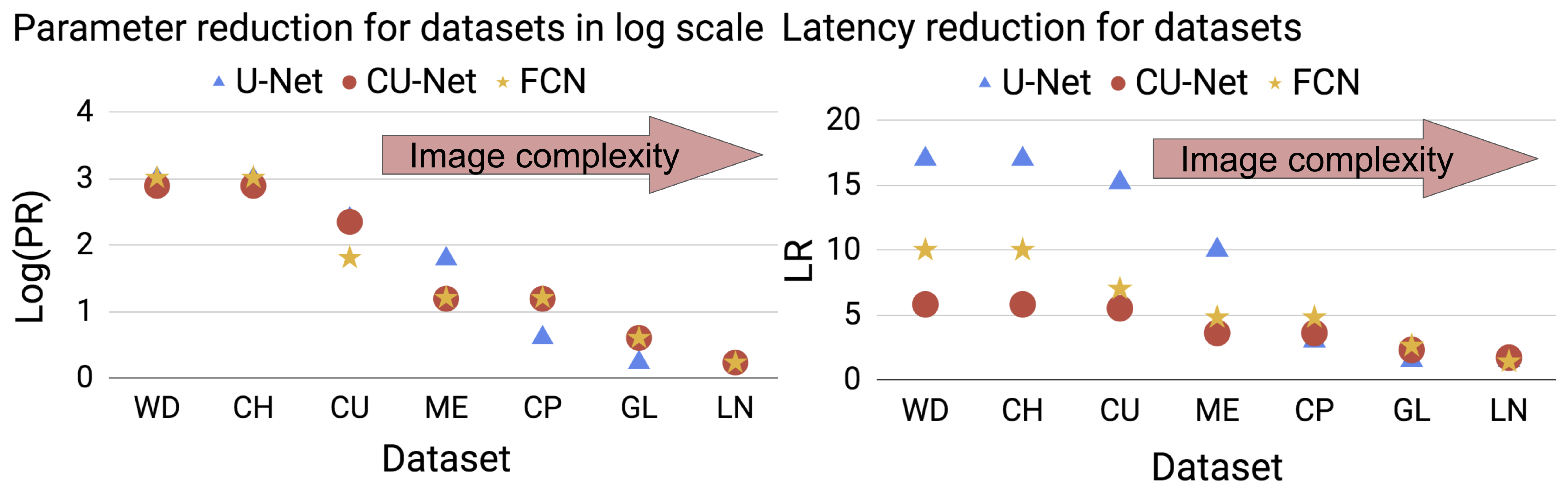}
    \vspace*{-0.2in}
\caption{Trainable parameter and inference latency reduction achieved (on test case 1) for various datasets, arranged along the $x$-axis in increasing image complexity.}
\label{fig:compression}
    \vspace*{-0.2in}
\end{figure}

\textbf{Test case 1 (accuracy-guided least memory usage).} We consider an  example constraint of $F1_{compressed} \geq 95\%F1_{base}$. The $\Delta \log \theta$ is estimated using $\lambda$ and $\delta$ and complexity (Table \ref{tab:data}). Using the ceiling $\alpha$ values, compressed networks are trained and analyzed. As shown in Table \ref{tab:evaluation}, a significant compression is achieved (best 113x for C2DH-U373 on U-net and least 3.5x for C2DL-PSC on CUMed) with much better accuracy compared to compression achieved using only \cite{squeezenet} or \cite{nvidiapruning}. To validate the effectiveness in estimating $\alpha$, we introduce a small reduction in $\alpha$ value ($\epsilon = \frac{1}{64}$, smallest possible keeping integer filters); the accuracy degrades below $95\%$ (Table \ref{tab:evaluation}, row CC-Net-case1-$\epsilon$). CC-Net compression does not show much improvement when pruned further, indicating few remaining ineffective filters. 

\textbf{Test case 2 (memory-constrained best possible accuracy).} We consider a disk space budget of 1 MB. Using ceiling of $\alpha = \sqrt{\theta^*/\theta}$, compressed networks are produced as shown in Table \ref{tab:evaluation}, whose accuracy satisfies the accuracy prediction made by our method (Fig.~\ref{fig:effectiveness}). 

\begin{table*}[thb!]
		\caption{Segmentation accuracy and network parameters on the C2DH-U373 and C2DL-PSC datasets.}
		\label{tab:evaluation}
		\centering
		\scriptsize
		\begin{tabular}{ c | c | c | c | c | c || c | c | c || c | c | c}
\hline
\multicolumn{3}{c|}{} & \multicolumn{3}{|c|}{U-Net \cite{unet}} & \multicolumn{3}{|c|}{CUMedVision \cite{cumednet}} & \multicolumn{3}{|c}{FCN \cite{fcn}} \\
\hline
\rule{0pt}{8pt} & Method & Dataset & F1 & IU & log(\#P) & F1 & IU & log(\#P) & F1 & IU & log(\#P)\\ \hline
& \multirow{2}{*}{\centering Base Network} & C2DH-U373 & 0.896 & 0.900 & \multirow{2}{*}{{\centering 7.492}} & 0.891 & 0.895 & \multirow{2}{*}{{\centering 6.887}} & 0.891 & 0.894 & \multirow{2}{*}{{\centering 7.552}} \\
 & & C2DL-PSC & 0.801 & 0.820 &  & 0.793 & 0.814 & & 0.755 & 0.788 & \\ \hline
\parbox[t]{3mm}{\multirow{14}{*}{\rotatebox[origin=c]{90}{Compressed Networks}}} & \multirow{2}{*}{{\centering Base Network + Squeeze \cite{squeezenet}}} & C2DH-U373 & 0.819 & 0.854 & \multirow{2}{*}{{\centering 7.049}} & 0.832 & 0.863 & \multirow{2}{*}{{\centering 6.669}} & 0.844 & 0.875 & \multirow{2}{*}{{\centering 7.369}} \\
 & & C2DL-PSC & 0.752 & 0.781 & & 0.751 & 0.781 & & 0.697 & 0.753 & \\ 
& \multirow{2}{*}{{\centering Base Network + Prune \cite{nvidiapruning}}} & C2DH-U373 & 0.858 & 0.867 & 7.491 & 0.848 & 0.861 & 6.886 & 0.809 & 0.837 & 7.551 \\
 & & C2DL-PSC & 0.749 & 0.785 & 7.491 & 0.744 & 0.768 & 6.886 & 0.691 & 0.738 & 7.552\\ 
& \multirow{2}{*}{{\centering CC-Net-\textbf{case1}}} & C2DH-U373 & \textbf{0.863} & \textbf{0.890} & \textbf{5.436} & \textbf{0.868} & \textbf{0.866} & \textbf{5.378}  & \textbf{0.880} & \textbf{0.885} & \textbf{5.939} \\
 & & C2DL-PSC & \textbf{0.775} & \textbf{0.818} & \textbf{6.640} & \textbf{0.763} & \textbf{0.794} & \textbf{6.341} & \textbf{0.720} & \textbf{0.766} & \textbf{6.949} \\ 
& \multirow{2}{*}{{\centering CC-Net-case1 + Squeeze}} & C2DH-U373 & 0.806 & 0.840 & 5.243 & 0.820 & 0.853 & 5.245 & 0.824 & 0.860 & 5.915 \\
 & & C2DL-PSC & 0.681 & 0.735 & 6.197 & 0.629 & 0.705 & 6.176 & 0.663 & 0.728 & 6.786\\ 
& \multirow{2}{*}{{\centering CC-Net-case1 + Prune}} & C2DH-U373 & 0.834 & 0.847 & 5.435 & 0.834 & 0.847 & 5.377 & 0.830 & 0.843 & 5.938 \\
 & & C2DL-PSC & 0.772 & 0.800 & 6.639 & 0.750 & 0.786 & 6.341 & 0.678 & 0.730 & 6.949 \\  
& \multirow{2}{*}{{\centering CC-Net-case1-$ \epsilon$}} & C2DH-U373 & 0.841 & 0.872 & 5.277 & 0.816 & 0.849 & 5.297 & 0.817 & 0.844 & 5.847 \\
& & C2DL-PSC & 0.751 & 0.781 & 6.603 & 0.759 & 0.785 & 6.315 & 0.713 & 0.742 & 6.922\\  
& \multirow{2}{*}{{\centering CC-Net-\textbf{case2}}} & C2DH-U373 & 0.832 & 0.863 & 5.097 & 0.807 & 0.837 & 5.097 & 0.803 & 0.834 & 5.097 \\
& & C2DL-PSC & 0.698 & 0.745 & 5.097 & 0.711 & 0.743 & 5.097 & 0.644	& 0.719 & 5.097\\
\hline
\end{tabular}
  \vspace*{-0.1in}
\end{table*}

The overall reduction (R = $\frac{base}{compressed}$) in trainable parameters (PR) and evaluation latency (LR) for all 7 datasets (for test case 1) is plotted in Fig.~\ref{fig:compression}. Larger complexity results in less compression, indicating a higher requirement in trainable parameters for extracting features. CC-Net achieves parameter and latency reduction in the range of $1000x$ to $ 2x$ and $17x$ to $1.5x$ for different datasets.

\begin{table}[thb!]
\caption{Training time consideration (test case 1)}
\label{tab:reduction}
\scriptsize
\begin{center}
\begin{tabular}{ c  c  c  c  c }
\hline
\rule{0pt}{6pt}Approach & Dataset & Pre-training & Training & Post-training \\ \hline
\multirow{2}{*}{\centering U-Net+\cite{nvidiapruning}} & C2DH-U373 & \multirow{2}{*}{\centering -} & 10781 ms & 160 \\
 &  C2DL-PSC & & 2348 ms & 30 \\
\multirow{2}{*}{\centering Ours (new)} & C2DH-U373 & \multirow{2}{*}{\centering O} & 4786 ms & - \\
 & C2DL-PSC &  & 1282 ms & - \\
\multirow{2}{*}{\centering Ours (existing)}  & C2DH-U373 & \multirow{2}{*}{\centering Negligible} & 4786 ms & - \\ 
 & C2DL-PSC & & 1282 ms & - \\
\hline
\end{tabular}
\end{center}
\vspace*{-0.2in}
\end{table}

Table \ref{tab:reduction} shows training time for \cite{nvidiapruning} and CC-Net on U-Net for test case 1 (on P100 GPU). Per epoch training time (in ms) is provided along with number of pruning epochs (column Post-training). We have used fewer fine-tuning iterations per pruning epoch, however, pruning is expensive and can exceed original network training by a factor of 3 \cite{deepcompression,nvidiapruning}.
One time 
$\lambda$ determination (`O' in Table \ref{tab:reduction}) for any CNN is a bottleneck for CC-Net. Yet, after this process, significant reduction in training time can be achieved for any dataset, trained on the same network. We consider `O' can be computed under 2x training time of base architecture, with a sufficient degree of accuracy,  using 2 datasets with two $\alpha$ points ($\alpha \in \{0.25,0.03125\}$.

\section{Conclusions}
In this paper, 
we presented a new image complexity-guided network compression scheme, CC-Net, for biomedical image segmentation. Instead of compressing CNNs after training, we focused on pre-training network size reduction, exploiting image complexity of the training data. 
Our method is effective in quickly generating compressed networks with target accuracy, outperforming state-of-the-art network compression methods. Our scheme accommodates practical applied design constraints for compressing CNNs for biomedical image segmentation. 

\section{Acknowledgement}
This work was supported in part by the National Science Foundation under Grants CNS-1629914, CCF-1640081, and CCF-1617735, and by the Nanoelectronics Research Corporation, a wholly-owned subsidiary of the Semiconductor Research Corporation, through Extremely Energy Efficient Collective Electronics, an SRC-NRI Nanoelectronics Research Initiative under Research Task ID 2698.004 and 2698.005.

\bibliographystyle{IEEEbib}
\balance
\bibliography{strings,refs}

\end{document}